\documentclass[
]{ceurart}
\usepackage{dirtytalk}
\usepackage{comment}
\begin{document}

\copyrightyear{2021}
\copyrightclause{Copyright for this paper by its authors.
  Use permitted under Creative Commons License Attribution 4.0
  International (CC BY 4.0).}

\conference{FIRE 21: Forum for Information Retrieval Evaluation, December 13–17, 2021, India}

\title{Overview of the Shared Task on Fake News Detection in Urdu at FIRE 2021}

\author[1]{Maaz  Amjad}[%
email=maazamjad@phystech.edu,
url=https://nlp.cic.ipn.mx/maazamjad/,
]
\address[1]{Instituto Politécnico Nacional (IPN), Center for Computing Research (CIC), Mexico }

\author[1]{Sabur Butt}[%
email= sabur@nlp.cic.ipn.mx,
]

\author[3]{Hamza Imam Amjad}[%
email=hamzaimamamjad@phystech.edu,
]

\author[2]{Alisa Zhila}[%
email=alisa.zhila@ronininstitute.org,
]

\author[1]{Grigori Sidorov}[%
email=sidorov@cic.ipn.mx,
]

\address[2]{Ronin Institute for Independent Scholarship, United States}

\address[3]{Moscow Institute of Physics and Technology, Russia}

\author[1]{Alexander Gelbukh}[%
email=gelbukh@gelbukh.com,
]

\begin{abstract}
Automatic detection of fake news is a highly important task in the contemporary world. This study reports the 2\textsuperscript{nd} shared task called UrduFake@FIRE2021 on identifying fake news detection in Urdu. The goal of the shared task is to motivate the community to come up with efficient methods for solving this vital problem, particularly for the Urdu language. The task is posed as a binary classification problem to label a given news article as a real or a fake news article. The organizers provide a dataset comprising news in five domains: (i) Health, (ii) Sports, (iii) Showbiz, (iv) Technology, and (v) Business, split into training and testing sets. The training set contains 1300 annotated news articles ---750 real news, 550 fake news, while the testing set contains 300 news articles ---200 real, 100 fake news. 34 teams from 7 different countries (China, Egypt, Israel, India, Mexico, Pakistan, and UAE) registered to participate in the UrduFake@FIRE2021 shared task. Out of those, 18 teams submitted their experimental results and 11 of those submitted their technical reports, which is substantially higher compared to the UrduFake shared task in 2020 when only 6 teams submitted their technical reports. The technical reports submitted by the participants demonstrated different data representation techniques ranging from count-based BoW features to word vector embeddings as well as the use of numerous machine learning algorithms ranging from traditional SVM to various neural network architectures including Transformers such as BERT and RoBERTa. In this year's competition, the best performing system obtained an F1-macro score of 0.679, which is lower than the past year's best result of 0.907 F1-macro. Admittedly, while training sets from the past and the current years overlap to a large extent, the testing set provided this year is completely different.

\end{abstract}


\maketitle

\section{Introduction}
The proliferation of social media brought in various forms of cybercrime that urgently need automatic solution for the safety of people online and beyond \cite{sexism_sabur, ashraf_2020, noman_ashraf_2020}. Among these problems, fake news dissemination is a critical problem that spreads in the form of advertisements, posts, news articles and others. It is an outstanding threat to journalism, democracy, and freedom of expression that negatively affects trust between the media outlets and the users. The socio-political impact of fake news can be observed with the incidents such as 2016 United States presidential elections. Post election studies showed~\cite{ali2021post, grinberg2019fake} various occasions of fake news spiking on social media with content emphasising nonexistent cause–effect relationship aggravating the division between the political groups. 
Behavioural studies~\cite{anderson2016social, bond201261} showed the effect that exposure to fake news has on political and social issues through randomized controlled experiments. The results established that fake news can cause a change in views and behaviour regarding topics of broad domain including politics. Hence, the status quo of fake news needs immediate attention and robust solutions.

Natural language processing (NLP) researchers formulated the problem into subcategories of fake news such as satire~\cite{burfoot2009automatic, reganti2016modeling}, propaganda~\cite{zubiaga2017exploiting, 12}, deception~\cite{b0, feng2012syntactic}, fact cherry picking~\cite{asudeh2020detecting, hendricks2019alternative}, clickbaits~\cite{chakraborty2016stop, chen2015misleading, potthast2016clickbait}, hyperpartisanship~\cite{14, jiang2019team}, and claim ``check-worthiness'' for potentially untruthful facts~\cite{hassan2017toward, hansen2019neural, CheckThat:ECIR2021, clef-checkthat:2021:LNCS, clef-checkthat:2021:task3, thorne2018fever}. Each subcategory has distinct features and solutions to achieve desirable results. Fake news becomes a very challenging problem to control because of the Velocity, Volume, Variety, and Time Latency of its spread~\cite{zhou2018fake}. The community behind the fake news content marches the spread at a pace which becomes higher than the real news dissemination itself.

This paper describes the UrduFake@FIRE2021 shared task and its results. The task invited the participants to tackle the problem of automatic fake news detection in Urdu in Nastal\'{i}q script . The problem is shaped into a binary classification problem in which news articles from various sources including such news outlets as BBC Urdu News, CNN Urdu, Express-News, Jung News, Naway Waqat, and others, are offered for classification as fake or real. During the active competition phase the ground truth annotations for the testing set were hidden from the participants, while the training set was provided with the corresponding ground truth annotations. After the end of the competition, the both parts of the dataset were made publicly available along with the corresponding ground truth annotations at the CICLing 2021 UrduFake track at FIRE 2021 shared task  homesite~\footnote{\url{https://www.urdufake2021.cicling.org/home}}. This year's track is the continuation of CICLing 2020 UrduFake track at FIRE 2020~\cite{amjad2020urdufake, 6} with the core difference being the size of the offered dataset. The training data 
has increased
to facilitate a wider range of neural network and particularly deep learning studies and to get more insightful information from data analysis. In the shared task the participating teams were requested to submit only their top 3 different runs, among which the best run was considered for submission of the technical report paper describing the approach. 

The paper is structured as follows. An overview of previous relevant research can be found in Section~\ref{sec:lit}.
We provide the task description in Section~\ref{sec:task}  
and explain in detail the data collection and annotation procedure in Section~\ref{sec:dataset}.
Training and testing set splits and statistics are outlined in Section~\ref{sec:traintest}.
Sections~\ref{sec:eval} and ~\ref{sec:baselines} describe the choice of evaluation metrics and baselines correspondingly. A high level overview and comparison of the solutions and approaches submitted by the participants is provided in Section~\ref{sec:overview} along with the final results summarized in Section Sections~\ref{sec:results}. 
A brief summary of the UrduFake@FIRE2021 track can be found in a separate publication~\cite{amjad2021urdufake}.

\section{Importance of Fake News Detection in Urdu}
\label{sec:importance}

Urdu is the national language of Pakistan and has more than 230 million~\footnote{\url{https://www.statista.com/statistics/266808/the-most-spoken-languages-worldwide/}} speakers worldwide. Many of these speakers carry out their written communication in the Nastal\'{i}q script. Urdu is commonly written in the Nastal\'{i}q script, while the  Devanagari script is commonly used for Hindi. However, due to cultural and geographical proximity, Devanagari may be also used for writing in Urdu. This creates a situation of \textit{digraphia} for the Urdu language when two scripts are used for writing in a language. Apart from this commonality, Urdu has other structural similarities with Hindi and other South Asian languages~\cite{adeeba2011experiences}. The emergence of Urdu came in the form of tribal movement which resulted in the merging of morphological and syntactic structures of Arabic, Persian, Turkish, Sanskrit, and recently English in the conversational usage. Due to the mixture of various languages, Urdu has more complexity than the other existing languages and, consequently, requires more careful processing. 

South Asia has been suffering from numerous instances of fake news affecting its political, social, and economic situation. For example, Dr. Shahid Masood~\footnote{\url{ https://www.globalvillagespace.com/dr-shahid-masoods-claims-about-zainabs-murderer-prove-false/ }} who works as a TV anchor in Pakistan, was exiled and tortured for spreading false information about a child rape case. 
Another case of fake news in India was reported in the Washington 
Post~\footnote{\url{https://tinyurl.com/ynhsudnx}},
where many innocent people died because of a child trafficking report. 

These severe consequences of fake news reporting surge the urge for high quality automation of fake news detection in Urdu. Given that despite the numerous speakers Urdu is still a low/medium resourced language, we strive for providing larger annotated datasets and incentivize the community to develop state-of-the-art solutions for early detection of fake news.

\section{Literature Review}
\label{sec:lit}

Contemporary fake news is not solely produced by humans, but can also be generated through bots~\cite{botsdetection}. These bots replicate human behaviour and are created for the purpose of spamming, spreading rumours and misinformation on various social media platforms. Social context~\cite{zhou2018fake} has been one of the key indicators to differentiate between fake and real news patterns. Researcher have dealt with the fake news problem with the aid of a wide range of feature based approaches~\cite{zuckerman1981verbal, nickerson1998confirmation, deutsch1955study, loewenstein1994psychology, ashforth1989social} including features such as engagement, user attributes, stylistic features, linguistic features, and personality based features. 

Earlier solutions~\cite{zhou2018fake} in fake news detection used fact checking with the aid of experts, however, the solution was time consuming and labor-cost intensive. Hence, NLP experts moved on to finding automatic solutions based on machine learning and deep learning algorithms~\cite{ashraf2021cic, Ekphrasis, 19, roy2018deep}. Studies have found unique emotional language cues~\cite{21} and emotional pattern~\cite{20} differences between real and fake news. Among the supervised machine learning techniques~\cite{ashraf2021cic, biyani20168, wu2015false}, we have seen Random Forest (RF), Support Vector Machine (SVM), and Decision Trees repeatedly used for fake news detection. Other research have used neural network ensembles combining various neural network architectures. Thus, Roy et al.~\cite{roy2018deep} fed article representations provided by CNN and Bi-LSTM models into MLP for the final classification which allowed for considering more contextual information. Yet another approach towards identifying fake news is looking at the news sources instead of the text content in the article, as news sources can provide valuable insights~\cite{4}. 

The dataset created for fake news identification mostly rely on social media platforms and news outlets. The majority of the existing datasets are available in English~\cite{6, zhou2018fake}. Recently, datasets and studies on various subcategories of fake news appeared in other languages: Persian \cite{10}, Spanish~\cite{4, 7}, Arabic~\cite{5,9}, German \cite{8}, Bangla~\cite{12}, Dutch \cite{14}, Italian~\cite{15}, Portuguese~\cite{13}, Urdu~\cite{a}, and Hindi~\cite{16}. 

Some of the online challenges to improve automatic fake news systems include Fake News Challenge~\footnote{\url{http://www.fakenewschallenge.org/}}, multiple fake news detection competitions on Kaggle~\footnote{\url{https://www.kaggle.com/c/fake-news/data}, \url{https://www.kaggle.com/c/fakenewskdd2020}} as well as shared task tracks organized by the academic community: PAN 2020~\cite{4}, RumourEval Task 8 of SemEval 2017 for English~\cite{17}, RumourEval Task 7 of SemEval-2019 for English~\cite{18}, and others.

\section{Task Description}
\label{sec:task}

This task is aimed to motivate the community to come up with methods and systems for automatic fake news detection in Urdu language by providing an annotated dataset with a train/test split and competitive settings. The challenge is posed as a binary classification task where participants are to train their classifiers on the provided training part of the dataset and to submit the labels, either fake or real, for each news article from the testing set, the ground truth annotations for the latter being hidden from the participants. Organizers compute the evaluation metrics for each submission by comparing the submitted labels to the ground truth annotations.   

The motivations of this shared task is to investigate whether and to which extent the textual content alone can be grounds for fake news detection and examine the efficiency of machine learning algorithms in identifying fake news articles written in Urdu in the Nastal\'{i}q script.

Here, a fake news article and fake news detection are defined as follows:

\begin{itemize}
\item {\verb|Fake News|}: A news article that contains factually incorrect information with the intention to deceive a reader and to make the reader believe that it is factually correct.

 \item {\verb|Fake News Detection|}: Suppose that $n$ is a news article (without annotation) and $n$  $\in$ $N$, where $N$ is the total number of news articles. A fake news detection is a process in which an algorithm calculates the likelihood of whether a given news article $n$ is a fake news article by assigning a value between 0 and 1. In mathematical terms, this can be described as $S(n)$ $\in$ [0, 1]. In other words, if $S(\widehat{n})$ > $S(n)$, this indicates that the $\widehat{n}$ new article has a higher chances to be fake news than the $n$ news article. Also, it is important to define a threshold. The threshold $\beta$ is a hyperparameter cut value selected by the algorithm developers such that if the algorithm assigns an equal or higher value to a news article as compared to the threshold, then the news article will be tagged as fake. A threshold $\beta$ can be defined so that the prediction function $F(n)$: $n$ $\rightarrow$  \{\textit{not fake}, \textit{fake}\} is:  
 
\begin{align*}
F(N) &=  \begin{cases}
fake, \,\,\,\,\,\,if \,\,\,\,\ S(n) >= \beta), \\
not fake, \,\,\,\,\,\,\,\,\  \textrm{otherwise}.
\end{cases}
\end{align*}
\end{itemize}

More elaborated definition of fake news is provided in our previous work \cite{a}. 

\section{Dataset Collection and Annotation}
\label{sec:dataset}

This section gives an outline of the dataset created for the UrduFake shared task at FIRE 2021. Our previous research \cite{a} reported the first version of this dataset, called “Bend The Truth” that contained 500 real news and 400 corresponding fake news. A new training dataset and test dataset data was acquired using the dataset collection and annotation guidelines presented in our previous research \cite{a}. The dataset presented in this shared task is publicly available and can be used for research objectives \footnote{https://github.com/MaazAmjad/Urdu-Fake-news-detection-FIRE2021}.

The training dataset was released on April 30, 2021 \footnote{https://www.urdufake2021.cicling.org/home}. It is important to mention that the training dataset used in 2021 UrduFake task comprised 1300 news article. This dataset was made up by combining the training dataset, which we presented in our previous research \cite{a} ``Bend The Truth" and testing dataset collected for UrduFake 2020 shared task. The training dataset contained 750 real news articles and 550 fake news articles. we presented a new test dataset that contained 200 real news and 100 fake news articles collected from January 2021 to August 2021 to test the proposed systems.

A crowdsourcing technique was used to collect the fake news articles. In other words, the fake news were composed by hiring professional journalists who deliberately wrote fake news of the corresponding real news. The journalists were provided a set of instructions to follow while writing fake news articles. This dataset contains five domains of the news: (i) Business, (ii) Health, (iii) Sports, (iv) Showbiz (entertainment), and (v) Technology.

\subsection{Procedure for Dataset Annotation }

\begin{figure}[htb!]
    \centering
    \includegraphics[width=0.5\textwidth]{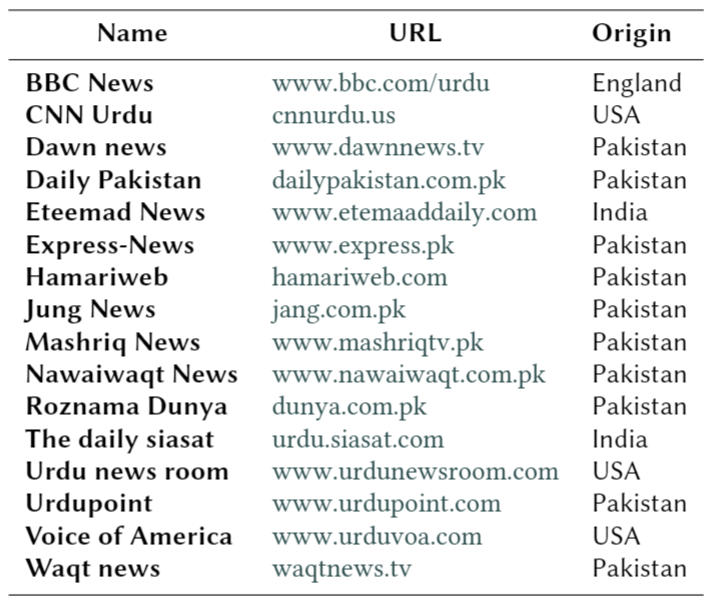}
    \caption{Legitimate websites}
    \label{fig:my_label}
\end{figure}

All the news articles were labelled into two two types of news: (i) real news article, and (ii) fake news article. Different techniques were used to annotate and assemble real and fake news. This dataset can be used for future research using supervised machine learning and deep learning techniques. Figure~\ref{fig:my_label} shows the list of news organizations used to crawl news articles. 

\subsubsection{Real News Collection and Annotation}

To assemble real news articles, various traditional news media mainstream were used to crawl news manually. Manual procedures were followed to annotate a news article using the underline guidelines, the news would label as real news. The news organizations used to gather news items for annotation are presented in Figure~\ref{fig:my_label} and all the news were manually crawled. The following guidelines were used to annotate a news item as a real news:

\begin{enumerate}
\item The news article was labeled as real news if the news meets the following criteria:
\begin{itemize}

\item That news article is published by a credible newspaper or a prominent news media agency.
\item The integrity of that news article can be verified by other credible newspaper agencies. This was an important point to do fact-checking. For example, manual source verification was performed to check place of the event, image, date of the news and whether the provided information in the news article matched with the same news article but published by other newspaper or news agency as well.
\item Incongruity between news titles and its content was also confirmed to ensure that a news article has a correlation between the news headline and the body text. We read the complete news articles to check the incongruity between news titles and the body text. 
\end{itemize}
\end{enumerate}

It is important to highlight that a news article was removed If it did not fulfil one of the aforementioned criteria. Different news articles contained different words length. For example, CNN publish news articles that contains between 200-300 words. On the other hand, a news article published by BBC Urdu news typically contains on average 1500 words. Therefore, the real news articles contains heterogeneous length of words. This is how all the real news articles were collected and annotated. 

\subsubsection{Professional Crowdsourcing of Fake News}
    
To obtain fake news, the services of professional journalist were used who work in different news organizations in Pakistan. We hired professional columnist because they are expertise in writing news articles, and use different journalists techniques to make the news interesting to hook and and their written fake news can easily trick the reader. The real news articles were provided to the journalists and they were asked to write fake news corresponding to the real news. In other words, if a real news contains story about football, the correspond fake news article should also contain similar story but with fabricated information.

We used professional “crowdsourcing” for collecting fake news and the reasons are described as follows:

\begin{enumerate}
\item The news articles analysis with manual procedures for verification through web scraping approach was unfeasible. This is due to the facet that it is extremely challenging task to find the corresponding fake news of a real news article. 

\item No online service in Urdu language is available for news fact-checking. Unlike English, the news fact-checking is manually performed in Urdu. 

\end{enumerate}

This dataset contains news of five domains: (i) business, (ii) education, (iii) sports, (iv) showbiz (entertainment), and (v) technology. The journalists expertise was taken into account to ensure that the fake news corresponding to the real news is written by the domain expert. The journalists were asked to keep the same length of the news (fake news article should have the same words length as real news). In addition, we also instructed journalists to mitigate defined patterns so that the undesirable clues should not be induced to classify news articles. Therefore, journalists’ expertise were used to collected all the fake news articles.


\subsection{Training and Testing Split}		
\label{sec:traintest}

\subsubsection{Training and Validation Set}

The training set contained 1300 news articles, in which 750 news articles were annotated as real, and 550 news articles were annotated as fake news article. The training set and the testing set contained five types of news: (i) Business, (ii) Health, (iii) Showbiz (entertainment), (iv) Sports, and (v) Technology. Participants were allowed to use of the training set for validation, development, and parameter tuning.  The training dataset made up by combining the training dataset, which we presented in our previous research \cite{a} ``Bend The Truth" and the testing dataset collected for UrduFake 2020 shared task.

\subsubsection{Test dataset}
The new test set was introduced that contained 200 real news and 100 fake news articles collected from January 2021 to August 2021. The test set was presented without the ground truth labels so that all the participants could evaluate and test the performance of their proposed systems. To evaluate and compare the performance of the classifiers submitted by the participants, the organizers used the truth labels of the test set. It is worth mentioning that the participants were unaware of the distributions of real and fake news in the test set.

\subsection{Dataset Statistics}

In this shared task, we divided the dataset into two parts: (i) training set, and (ii) testing set. Initially, the training set was released so that the participants can train their classification models. Then, the test set was released so that the participants can predict the labels of whether a given news is real or fake.  Table \ref{tab:Corpus_distribution} describes the corpus distribution of the news articles by topics for the training and testing sets.

\begin{table}[htb!]
	\caption{Domain Distribution in Train and Test subsets}
	\label{tab:Corpus_distribution}
	  \centering
		\begin{tabular}{ lcccc }
			\hline\noalign{\smallskip}			
			\multirow{2}{*}{\textbf{Domain}}& \multicolumn{2}{c}{\textbf{Train}} & \multicolumn{2}{c}{\textbf{Test}}\\
			& \textbf{real} & \textbf{fake}& \textbf{real} & \textbf{fake} \\
			\noalign{\smallskip}\hline\noalign{\smallskip}
			\textbf{Business} & 150& 80 & 40	& 20\\
			\textbf{Health} & 150& 130 & 40	& 20\\		
			\textbf{Showbiz}& 150& 130 & 40	& 20\\
			\textbf{Sports} & 150& 80 & 40	& 20\\
			\textbf{Technology} & 150& 130 & 40	& 20\\
			\noalign{\smallskip}\hline\noalign{\smallskip}
			\textbf{Totals} &\textbf{750}&\textbf{550}& \textbf{200}	& \textbf{100}\\		
			\noalign{\smallskip}\hline
		\end{tabular}
\end{table} 

\section{Evaluation Metrics}
\label{sec:eval}

This is a binary classification task in which the task is to classify a news article as fake or real. All the participating teams were allowed to submit up to 3 different runs, i.e., labels for the testing set generated by their proposed classifiers. The ground truth annotations were used to compare the labels predicted by the participants' classifiers. We used the evaluation metrics commonly used to measure the performance of binary classification on imbalanced datasets: two sets of \textit{Precision} (P), \textit{Recall} (R), and \textit{F1} score, one for the ``real'' class treated as a target class and the other for the ``fake'' class; the inter-class metrics \textit{Accuracy} and \textit{F1-macro}. The macro-averaged F1-macro, which is the average of F1\textsubscript{real} and F1\textsubscript{fake}, was also calculated to accommodate the dataset skew towards the real class. As detection of both classes (real and fake) is equally important, this is why we evaluated performance against both classes.

\section{Baselines}
\label{sec:baselines}

To introduce a baseline, we used the bag of words (BoW) approach. 
We used a combination of character, word, and function word bi-grams with TF-IDF weighting scheme for text representation. Function words are similar to stopwords, for more elaborated definition and list we suggest to refer to \cite{3}. Decision Tree was selected as a classifier, which achieved surprisingly good results compared to other traditional ML classifiers on our trial runs. 
d
In the trial runs, five weighting schemes (tf-idf, log-ent, norm, binary, relative frequency) \cite{a} were used for the experiments along with different machine learning classifiers such Decision Tree, Random Forest, Logistic Regression, AdaBoost,  SVM, and Naive Bayes. We tried different $n$-grams, $n=\{1, ..., 7\}$. We noticed that the classifiers started to obtain insignificant results when $n$ = 5 or higher. Finally, the Decision Tree algorithm outperformed other classifiers in identifying fake news. The baseline code is publicaly available~\footnote{ https://github.com/MaazAmjad/Urdu-Fake-news-detection-FIRE2021}.

\section{Overview of the Submitted Approaches}
\label{sec:overview}

This section briefly overviews the methods applied in the competition by the teams. In total 34 teams registered for the competition, and 18 teams submitted experimental results on a test dataset. We report the findings of 11 teams who submitted their methodologies in the form of technical report papers. The registered participants were from the countries where Urdu language has presence or cause interest: Pakistan, India, United Arab Emirates, Israel, and Egypt. Table~\ref{dic} shows the approaches used by the teams and table~\ref{tab:overlapping2} tells the best run scores achieved through those methods. 

\begin{enumerate}

\item \textbf{Nayel: } The best performing model used the linear classifier function from the \textsc{scikit-learn} package {\verb|sklearn.linear_model.SGDClassifier|} that by default fits a linear SVM classifier with Stochastic Gradient Descent (SGD) optimization algorithm. The team trained it on word token tri-gram features weighted with TF-IDF scheme. The model uses tokens without any preprocessing, which increases the number of features.

\item \textbf{Abdullah-Khurem: }  The team experimented with neural network techniques: Convolutional Neural Network (CNN), 
Recurrent Neural Network (RNN) and textCNN, for fake news detection in Urdu. The final submission used textCNN with TF-IDF features which and ranked second in the competition.

\item \textbf{Hammad-Khurem: } The methodology used no pro-processing and proposed a voting-based approach with a majority voting ensemble of boosting-based ML classifiers: AdaBoost, LightGBM and XGBoost. The proposed approach employed BoW features. 

\item \textbf{Muhammad Homayoun: } The participant reported results using Convolution Neural Network (CNN) with four input channels. Before classification, the data was pre-processed by removing diacritic, normalization, stopword removal and lemmatization. The best results submitted used character level sequences (n-grams) for text representation.

\item \textbf{Snehaan Bhawal: }The transformer methods (MuRIL, BERT) gave the best results with no pro-processing. Multilingual Representations for Indian Languages (MuRIL) was submitted to the competition as the final submission and slightly outranked the non-specialized BERT.

\item \textbf{MUCIC: }The participants used three feature selection algorithms (Chi-square, Mutual Information Gain (MIG), and f\_classif) to choose the best features from the word and character n-grams. The intersection of selected features was passed into an ensemble of ML classifiers (Linear SVM (LSVM), LR, MLP, XGB, and RF) with soft voting and feature selection to achieve the best results. 

\item \textbf{SOA NLP: }The submitted method used character level uni, bi and tri-gram TF-IDF features as an input to dense neural network (DNN). The best results used a learning rate of 0.001, a dropout rate of 0.3, a batch size of 16, Adam as an optimizer and binary cross-entropy as a loss function with 100 epoch training.

\item \textbf{Dinamore\&Elyasafdi\_SVC: }The team used classical machine learning algorithms: SVM, Random Forest (RF), and Logistic Regression (LR). They used character tri-gram features with only one pre-processing step of lowercasing all letters.

\item \textbf{MUCS: } In the pre-processing stage the participants removed non-relevant characters, stopwords and punctuation. They used pre-trained Urdu word embeddings from fastText and TF-IDF of words as well as character n-grams as features. Similar to team MUCIC, an  ensemble of ML classifiers (RF, MLP, AdaBoost, and GraidentBoost) were used with soft voting to achieve the highest F1 macro. 

\item \textbf{Iqra Ameer: }This is another study that used BERT-base model. The best results were reported using both the training and validation set for training of the model.

\item \textbf{Sakshi Kalra: } The best team runs used an ensemble of various transformer methods (RoBERTa, XLM-RoBERTa and Multilingual BERT) as well as a single specialized transformer \textsc{RoBERTa-urdu-small}. The text input was normalized. Interestingly, the best performing method on the test set turned out to be  \textsc{RoBERTa-urdu-small} which exceeded the three-transformer ensemble method (XLM-RoBERTa+Multilingual BERT+RoBERta).

\end{enumerate}

\begin{table}[!htb]
  \caption{Approaches used by the participating teams}
  \resizebox{\columnwidth}{!}
  {%
   \label{dic}
  \begin{tabular}{ccccc}
    \toprule
\textbf{System/Team Name} & \textbf{Text Representation} & \textbf{Feature Weighting Scheme} & \textbf{Classifying Algorithm} & \textbf{is NN-based?}\\
    \midrule
    
Nayel & tri-gram  &  TF-IDF  &  linear SVM with SGD & No \\
Abdullah-Khurem &  Word2Vec, GloVe, fastText &  TF-IDF     &  textCNN   & Yes \\
Hammad-Khurem & BoW  &  count \textit{(?)}   & ensemble XGBoost+LightGBM+AdaBoost    & No \\
Muhammad Homayoun  & char ${2,6}$-gram    &  \textit{N/A}   & CNN & Yes \\
Snehaan Bhawal  & transformer embeddings   &  \textit{N/A}   & MuRIL    & Yes \\
MUCIC & word- \&  char ${1,2}$-grams   & TF-IDF    & ensemble linSVM+LR+MLP+XGB+RF    & Yes (MLP) \& No \\
SOA NLP & char ${1,3}$-grams  & TF-IDF    & DNN  & Yes \\
Dinamore\&Elyasafdi\_SVC  & char 3-grams   &  TF-IDF   & SVM & No\\
MUCS & word fastText emb \& char ${2,3}$-grams  & TF-IDF for char-grams & ensemble MLP+AdaBoost+GraidentBoost+RF    & Yes (MLP) \& No \\
Iqra Ameer & transformer emb   &  \textit{N/A}   & BERT-base    & Yes \\
Sakshi Kalra & transformer emb   &  \textit{N/A}    & RoBERTa-urdu-small  & Yes \\

  \bottomrule
\end{tabular}
}
\end{table}

\section{Results and Discussion}
\label{sec:results}

Each team submitted three runs (proposed three different systems), and only the best run was considered for comparison. We calculated the results of all the submitted runs by each teams individually and only reported the results obtained by the best run. Table {\ref{tab:overlapping2}} shows the the results of the best run (among up to three submitted runs) submitted by the participating teams. We used F1-macro score to rank the participants systems. The aggregated statistics about the performance is presented in Table \ref{tab:overlapping33}.

It can be observed that only two systems outperformed the baseline, and all the other systems did not beat the F1-macro score of the baseline. The team Nayel obtained the the best results in terms of F1-macro, Accuracy,  P\textsubscript{fake} (precision) scores. The team Abdullah-Khurem obtained the the second best results in terms of F1-macro, Accuracy,  R\textsubscript{fake} (recall), P\textsubscript{real} (precision), and F1\textsubscript{fake} scores. Moreover, the baseline approach with the combination of char-word-function words bi-gram with tf-idf weighting scheme using Decision Tree classifier the third position in the shared task with the difference of 2.8\% from Nayel system and 1.2\% from Abdullah-Khurem system in F1-macro score. 

Table {\ref{tab:overlapping2}} presents the best results of the submitted systems.   

\begin{table}[!ht]
	\caption{Participants’ best run scores. }
	  \resizebox{\columnwidth}{!}{%
	\label{tab:overlapping2}
	  \centering
\begin{tabular}{lllllllllllll}
\rowcolor[HTML]{C2D69B} 
\cellcolor[HTML]{B8CCE4}                              & \cellcolor[HTML]{B8CCE4}                                      & \multicolumn{4}{c}{\cellcolor[HTML]{C2D69B}\textbf{Fake Class}}                              & \multicolumn{4}{c}{\cellcolor[HTML]{C2D69B}\textbf{Real Class}}                             & \multicolumn{2}{l}{\cellcolor[HTML]{C2D69B}}          &          \\
\rowcolor[HTML]{C2D69B} 
\multirow{-2}{*}{\cellcolor[HTML]{B8CCE4}\textbf{No}} & \multirow{-2}{*}{\cellcolor[HTML]{B8CCE4}\textbf{Team Names}} & Prec              & Recall            & \multicolumn{2}{l}{\cellcolor[HTML]{C2D69B}F1\_Fake} & Prec             & Recall            & \multicolumn{2}{l}{\cellcolor[HTML]{C2D69B}F1\_Real} & \multicolumn{2}{l}{\cellcolor[HTML]{C2D69B}\textbf{F1\_Macro}} & Accuracy \\
\cellcolor[HTML]{FFFFFF}1                             & Nayel                                                         & 0.754             & 0.400             & \multicolumn{2}{l}{0.522}                            & 0.757            & 0.935             & \multicolumn{2}{l}{0.836}                            & \multicolumn{2}{l}{0.679}                             & 0.756    \\
\cellcolor[HTML]{FFFFFF}2                             & Abdullah-Khurem                                               & 0.592             & 0.480             & \multicolumn{2}{l}{0.530}                            & 0.762            & 0.835             & \multicolumn{2}{l}{0.797}                            & \multicolumn{2}{l}{0.663}                             & 0.716    \\
\rowcolor[HTML]{FDE9D9}3                             &  Baseline                              & 0.584             & 0.450             & \multicolumn{2}{l}{0.508}                            & 0.753            & 0.840              & \multicolumn{2}{l}{0.794}                            & \multicolumn{2}{l}{0.651}                             & 0.710    \\
\cellcolor[HTML]{FFFFFF}4                             & Hammad-Khurem                                                 & 0.634             & 0.330             & \multicolumn{2}{l}{0.434}                            & 0.729            & 0.905             & \multicolumn{2}{l}{0.808}                            & \multicolumn{2}{l}{0.621}                             & 0.713    \\
\cellcolor[HTML]{FFFFFF}5                             & Muhammad Homayoun                                             & 0.480             & 0.490             & \multicolumn{2}{l}{0.485}                            & 0.742            & 0.735             & \multicolumn{2}{l}{0.738}                            & \multicolumn{2}{l}{0.611}                             & 0.653    \\
\cellcolor[HTML]{FFFFFF}6                             & Snehaan bhawal                                                & 0.960             & 0.240             & \multicolumn{2}{l}{0.384}                            & 0.723            & 0.995             & \multicolumn{2}{l}{0.837}                            & \multicolumn{2}{l}{0.610}                             & 0.743    \\
\cellcolor[HTML]{FFFFFF}7                             & MUCIC                                                         & 0.821             & 0.230             & \multicolumn{2}{l}{0.359}                            & 0.716            & 0.975             & \multicolumn{2}{l}{0.826}                            & \multicolumn{2}{l}{0.592}                             & 0.726    \\
8                                                    & SOA NLP                                   &    0.793          &     0.230          & \multicolumn{2}{l}{0.356}                            &      0.356       &     0.715         & \multicolumn{2}{l}{0.823}                            &  \multicolumn{2}{l}{0.590}                             &  0.590   \\                       
9                                                    & Dinamore \& Elyasafdi \_SVC                                   & 0.720             & 0.180              & \multicolumn{2}{l}{0.288}                            & 0.701            & 0.965             & \multicolumn{2}{l}{0.812}                            & \multicolumn{2}{l}{0.550}                             & 0.703    \\
10                                                    & MUCS                                                          & 0.850             & 0.170             & \multicolumn{2}{l}{0.283}                            & 0.703            & 0.985             & \multicolumn{2}{l}{0.820}                            & \multicolumn{2}{l}{0.552}                             & 0.713    \\

11                                                    & Iqra 	Ameer                                                   & 0.454             & 0.100             & \multicolumn{2}{l}{0.163}                            & 0.676            & 0.940             & \multicolumn{2}{l}{0.786}                            & \multicolumn{2}{l}{0.475}                             & 0.660    \\
12                                                    & Sakshi kalra                                                  & 0.266             & 0.120             & \multicolumn{2}{l}{0.165}                           & 0.654            & 0.835             & \multicolumn{2}{l}{0.734}                            & \multicolumn{2}{l}{0.449}                             & 0.596    \\

\end{tabular}}
\end{table}



Table {\ref{tab:overlapping33}} presents aggregated statistics of the submitted systems.


\begin{table}[!htb]
\centering
 \caption{Aggregated statistics of the submitted systems and the baseline.}
 	\label{tab:overlapping33}
\begin{tabular}{rrrrrrrrr}
\rowcolor[HTML]{FFFFFF} 
\multicolumn{1}{c}{\cellcolor[HTML]{FFFFFF}Stat.   metric} &
  \multicolumn{1}{c}{\cellcolor[HTML]{FFFFFF}P\textsubscript{fake}} &
  \multicolumn{1}{c}{\cellcolor[HTML]{FFFFFF}R\textsubscript{fake}} &
  \multicolumn{1}{c}{\cellcolor[HTML]{FFFFFF}F1\textsubscript{fake}} &
  \multicolumn{1}{c}{\cellcolor[HTML]{FFFFFF}P\textsubscript{real}} &
  \multicolumn{1}{c}{\cellcolor[HTML]{FFFFFF}R\textsubscript{real}} &
  \multicolumn{1}{c}{\cellcolor[HTML]{FFFFFF}F1\textsubscript{real}} &
  \multicolumn{1}{c}{\cellcolor[HTML]{FFFFFF}F1-macro} &
  \multicolumn{1}{c}{\cellcolor[HTML]{FFFFFF}Acc.} \\
\rowcolor[HTML]{F5F5F5} 
\textbf{mean}             & 0.596 & 0.294 & 0.348 & 0.679 & 0.837 & 0.757 & 0.553 & 0.658 \\
\rowcolor[HTML]{FFFFFF} 
\textbf{std}              & 0.201 & 0.171 & 0.136 & 0.105 & 0.201 & 0.137 & 0.093 & 0.106 \\
\rowcolor[HTML]{F5F5F5} 
\textbf{min}              & 0.262 & 0.070 & 0.115 & 0.356 & 0.155 & 0.228 & 0.296 & 0.303 \\
\rowcolor[HTML]{FFFFFF} 
\textbf{percentil   10\%} & 0.319 & 0.100 & 0.165 & 0.610 & 0.684 & 0.713 & 0.448 & 0.584 \\
\rowcolor[HTML]{F5F5F5} 
\textbf{percentil   25\%} & 0.462 & 0.145 & 0.229 & 0.680 & 0.802 & 0.761 & 0.490  & 0.646 \\
\rowcolor[HTML]{FFFFFF} 
\textbf{percentil   50\%} & 0.592 & 0.240 & 0.364 & 0.716 & 0.905 & 0.797 & 0.590  & 0.686 \\
\rowcolor[HTML]{F5F5F5} 
\textbf{percentil   75\%} & 0.737 & 0.430 & 0.451 & 0.726 & 0.970  & 0.816 & 0.61  & 0.713 \\
\rowcolor[HTML]{FFFFFF} 
\textbf{percentil   80\%} & 0.769 & 0.462 & 0.473 & 0.734 & 0.975 & 0.821 & 0.615 & 0.714 \\
\rowcolor[HTML]{F5F5F5} 
\textbf{percentil   90\%} & 0.826 & 0.506 & 0.510 & 0.753 & 0.981 & 0.828 & 0.653 & 0.729 \\
\rowcolor[HTML]{FFFFFF} 
\textbf{max}              & 0.960 & 0.600 & 0.530 & 0.762 & 0.995 & 0.837 & 0.679 & 0.756
\end{tabular}
\end{table}

\section{Conclusion}
Automatic fake news detection is an important task, especially in low resource languages. This research presents the second shared task (the first task was organized in 2020) in identifying fake news in Urdu namely the UrduFake 2021 track at FIRE 2021. A training and testing dataset was presented so that the participants could train and test their proposed systems. The dataset contained news in five domains (business, health, sports, showbiz, and  technology). All the real news were crawled from credible sources and manually annotated while the fake news were written by the professional journalists.  

In this shared task, thirty four teams from seven different countries registered and eighteen teams submitted their proposed systems (runs). The participants used different techniques ranging from the traditional feature-crafting and application of traditional ML algorithms to word representation through pre-trained embeddings to contextual representation and end-to-end neural network based methods. The approaches used included ensemble methods, CNN, and non-Urdu specialized Transformers (BERT, RoBERTa) as well as Urdu-specialized (MuRIL, RoBERTa-urdu-small)  . 

Team Nayel outperformed all the proposed systems by using the linear SVM optimized with Stochastic Gradient Descent and obtained F1-macro score of 0.67. This result reveals that classical feature-based models perform better compared to the contextual representation and large neural network algorithms. The characteristics of the dataset require further investigation to better explain this observation. 

 This shared task aims to attract and encourage researchers working in different NLP domains to address the automatic fake news detection task and help to mitigate the proliferation of fake content on the web. Moreover, this also offers a unique opportunity to explore the sufficiency of textual content modality alone and effectiveness of fusion methods. In addition, an annotated news dataset in Urdu is also provided to encourage more research to address the automatic fake news detection in Urdu language. 

\begin{acknowledgments}
This competition was organized with the support from the Mexican Government through the grant A1-S-
47854 of the CONACYT, Mexico and grants 20211784, 20211884, and 20211178 of the
Secretar\'{\i}a de Investigación y Posgrado of the Instituto Politécnico Nacional, Mexico.
\end{acknowledgments}

\bibliography{main}

\appendix

\end{document}